\documentclass{article}

\PassOptionsToPackage{numbers,compress}{natbib}

 \usepackage[preprint]{neurips_2026}

\usepackage[utf8]{inputenc} 
\usepackage[T1]{fontenc}    
\usepackage{graphicx}      
\usepackage{hyperref}       
\usepackage{url}            
\usepackage{booktabs}       
\usepackage{amsmath}        
\usepackage{amsfonts}       
\usepackage{amssymb}        
\usepackage{nicefrac}       
\usepackage{microtype}      
\usepackage{xcolor}         
\usepackage{enumitem}       
\usepackage{multirow}       

\newcommand{\metricci}[2]{\shortstack{#1\\{\scriptsize #2}}}

\title{AERIC\@: Anticipatory Hidden-State Monitoring for Implicit Harmful Dialogue}

\author{
  Jihyung Park \quad Saleh Afroogh  \quad Junfeng Jiao\thanks{Corresponding author.} \\
  The University of Texas at Austin \\
  \texttt{\{jihyung803, saleh.afroogh\}@utexas.edu} \\
  \texttt{jjiao@austin.utexas.edu}
}

\begin{document}

\raggedbottom%

\maketitle

\begin{abstract}
Current language models create two safety challenges: risk must be detected early enough to avoid exposing harmful continuation, and the harmfulness itself may be implicit rather than signaled by overtly toxic text. Existing response-level guards are strong at judging completed text, and native streaming guards move closer to token time, but both settings leave open whether a lightweight monitor can anticipate implicit harmful drift from the generator's own internal trajectory. We study anticipatory same-pass monitoring, where a safety monitor may read hidden states produced during ordinary decoding but may not invoke an additional forward pass through the base model. We introduce \textsc{AERIC}, a transfer-oriented hidden-state approach for implicit harmful dialogue that combines short-horizon hazard forecasting, support-sensitive suppression, and prompt-conditioned residual scoring under a same-pass exponential moving average decision rule. The default linear monitor contains only $387$ trainable head parameters. Against Qwen3Guard{-}Stream{-}4B on balanced benchmarks, \textsc{AERIC} improves AUROC from $0.6830$ to $0.7143$ on DiaSafety and from $0.8219$ to $0.8582$ on Harmful Advice. For prompt-level trigger benchmarks, we calibrate the \textsc{AERIC} threshold by a source-side safe-budget rule that maximizes trigger coverage while constraining the safe-trigger rate to at most $10\%$. Under that rule, trigger@$64$ reaches $0.6438$ and $0.4656$ on HarmBench DirectRequest and $0.6849$ and $0.7363$ on SocialHarmBench for Qwen and Gemma, respectively, withholding between $23.53$ and $41.86$ answer tokens on average. Same-pass deployment is also efficient: on a 63-prompt harmful-prompt fixed-generation benchmark aggregated over HarmBench DirectRequest and SocialHarmBench under Qwen3{-}8B, the monitor increases mean latency by only $2.34\%$, whereas Qwen3Guard{-}Stream{-}4B increases it by $79.40\%$. These results support a focused claim: explicit harmful supervision can transfer to implicit harm monitoring under a strict no-extra-forward-pass constraint, yielding a practical pre-exposure risk signal even against a strong native streaming guard, although intervention policy remains an open systems problem.
\end{abstract}

\section{Introduction}

Current LLM safety faces two partly distinct challenges. First, safeguards are typically strongest when judging fully observed text, whereas streaming deployment requires safety decisions over partial generations and therefore benefits from early intervention during decoding~\citep{li_judgment_2025,kavumba_predict_2026,xuan_shieldhead_2025}. Second, harmfulness is not always explicit: unsafe behavior can be covert, context-sensitive, or only implicit in the trajectory of a response, especially in dialogue and advice settings~\citep{mei_mitigating_2022,sun_safety_2022,qiu_benchmark_2023,wen_unveiling_2023,luettgau2025harmfuladvice, luettgau_people_2026}. Existing guards such as ShieldGemma and WildGuard are strong response-level moderators, but their default role is retrospective: they judge whether a prompt or a completed response is unsafe after enough text has already been produced. Even when moderation is moved closer to token time, detection often still depends on running an additional guard over growing prefixes, or on waiting until the generated text itself becomes sufficiently explicit to classify. In both cases, the system reacts after the unsafe trajectory has already become legible on the surface.

This paper studies a stricter objective: anticipatory same-pass monitoring. In streaming generation, safety decisions must be made over partial outputs rather than only after the full response is available, which pushes moderation toward fine-grained checks during decoding~\cite{zeng_root_2025,phan_think_2025}. But frequent safety checking can itself become a systems bottleneck: repeated intervention during generation increases inference cost and latency unless the monitoring signal is kept lightweight~\cite{yang_prefix_2025}. We therefore focus on a same-pass setting in which the monitor reads hidden states already produced by the generator during ordinary decoding, rather than repeatedly invoking a separate generative guard. This follows a broader shift toward deriving safety signals from internal representations to support real-time monitoring at lower cost~\cite{jiao_llm_2026}.

The harder challenge is that harmfulness is often implicit. In many dialogue settings, danger is not carried by a single explicit toxic phrase. A response may remain polite, supportive in tone, or locally reasonable while still drifting toward self-harm encouragement, unsafe medical guidance, exploitative escalation, or other context-dependent failures. This is especially true in dialogue safety and advice settings, where the same words can play very different roles depending on the prompt and the conversational context. Looking only at the visible prefix, or only at the model's top next-token preferences, often answers the wrong question. Those signals say what the model is likely to say next locally, but not whether the continuation has already begun to move toward an unsafe region of behavior~\citep{mei_mitigating_2022,sun_safety_2022,wen_unveiling_2023}.

Our approach starts from the observation that the model's hidden states can contain forward-looking information about continuation that is not fully recoverable from surface text alone. We introduce \textsc{AERIC}, Anticipatory Evidence and Residual Inference for Continuations, a same-pass hidden-state monitor for implicit harmful dialogue. \textsc{AERIC} combines three signals that correspond directly to the failure modes above. The first is a future-hazard head that predicts whether harmful continuation is about to begin within a short horizon. This is the mechanism that lets the monitor act before the harmful content becomes explicit on screen. The second is a support head that models counterevidence for safe, bounded, supportive, or de-escalatory continuation. This matters because emotionally intense language is not necessarily unsafe, and a monitor that only looks for hazard tends to overfire on difficult but still appropriate assistance. The third is a paired residual head that measures prompt-conditioned unsafe drift. Instead of asking only whether the current text looks risky in isolation, it asks whether the hidden-state trajectory has deviated toward an unsafe continuation relative to safe behavior for the same kind of prompt. The resulting raw score is deployed through a same-pass exponential moving average decision rule, which preserves prefix-measurability while stabilizing online triggers.

Across two generator families, this framing produces a consistent picture: \textsc{AERIC} reaches AUROC $0.7143$ on DiaSafety and $0.8582$ on Harmful Advice with \texttt{Qwen/Qwen3{-}8B}, and $0.7181$ and $0.8287$, respectively, with \texttt{google/gemma{-}4{-}E4B{-}it}. Qwen3Guard{-}Stream{-}4B is the strongest native streaming baseline in our comparison, yet \textsc{AERIC} remains above it in AUROC on both balanced targets. AUPRC margins are smaller and overlap in some comparisons, so we report both ranking metrics explicitly. On prompt-only harmful request suites evaluated under a source-side $10\%$ safe-trigger budget, the monitor remains actionable before exposure. On HarmBench DirectRequest it reaches trigger@64 $0.6438$ with Qwen and $0.4656$ with Gemma, while on SocialHarmBench it reaches $0.6849$ and $0.7363$, respectively. It also does so with low overhead. On a 63-prompt harmful-prompt fixed-generation aggregate over HarmBench DirectRequest and SocialHarmBench under \texttt{Qwen/Qwen3{-}8B}, the same-pass monitor increases mean latency by only $2.34\%$. The corresponding overheads are $79.40\%$ for Qwen3Guard{-}Stream{-}4B, $158.73\%$ for prefixized ShieldGemma{-}9B, and $216.74\%$ for prefixized WildGuard.

Existing guards are usually best at deciding whether harm is already present in completed or sufficiently explicit text, whereas our goal is to detect harmful continuation before exposure. Existing moderation signals also rely heavily on surface form, whereas our target is implicit harmful drift that depends on prompt context and discourse role. \textsc{AERIC} shows that explicit harmful supervision can be repurposed into an anticipatory same-pass monitor for implicit harmful dialogue by combining short-horizon hazard forecasting, support-sensitive suppression, prompt-conditioned residual scoring, and EMA-smoothed online triggering. This gives a practical pre-exposure signal under a strict no-extra-forward-pass constraint, even though the downstream intervention policy remains a separate systems problem. \footnote{To support reproducibility, code and evaluation scripts will be released with the camera-ready version.} 

\section{Related Work}

\paragraph{Implicit harmfulness.}
A central difficulty in language-model safety is that harmfulness is often implicit rather than explicit. Prior work argues that harmful text is not a single surface-level phenomenon and should be characterized along multiple dimensions, including context and pragmatic effect~\cite{rauh_characteristics_2022}. Work on covertly unsafe text and implicit toxicity further shows that dangerous or toxic meaning may not be reducible to obvious keywords, known surface patterns, slurs, or overtly toxic phrases~\cite{mei_mitigating_2022,hartvigsen_toxigen_2022,wen_unveiling_2023}. In dialogue and advice settings, unsafe meaning may depend less on a single toxic phrase than on conversational context, discourse role, or the likely direction of the continuation~\cite{sun_safety_2022,qiu_benchmark_2023}. DiaSafety~\cite{sun_safety_2022} is a representative benchmark for this setting, since safety judgments often depend on contextual interpretation rather than explicit wording alone. Harmful Advice~\cite{luettgau2025harmfuladvice, luettgau_people_2026} provides a complementary advice-oriented target in which harmfulness can remain locally subtle even when the overall recommendation is unsafe. Prior work on hidden-state probing further motivates our approach by showing that internal representations can reveal latent properties such as deception and future continuation structure before those properties are fully visible in surface text~\cite{azaria_internal_2023,pal_future_2023}. We build on that premise, but shift it to a transfer setting in which the goal is not simply to read out latent attributes, but to forecast unsafe continuation in implicit harmful dialogue from prefix hidden states.

\paragraph{LLM safeguards.}
A separate line of work studies safeguards more directly. ShieldGemma\footnote{\url{https://huggingface.co/google/shieldgemma-9b}}~\cite{zeng_shieldgemma_2024} and WildGuard\footnote{\url{https://huggingface.co/allenai/wildguard}}~\cite{han_wildguard_2024} are representative response-level guards that classify harmful prompts, harmful responses, and refusal behavior after text has already been produced. They are strong and practical baselines, and we compare against them directly, but their default operating mode is still retrospective. Recent work has therefore pushed moderation closer to token time. Qwen3Guard{-}Stream{-}4B\footnote{\url{https://huggingface.co/Qwen/Qwen3Guard-Stream-4B}}~\cite{zhao_qwen3guard_2025} is especially relevant in our setting because it is a native streaming guard, while NExT-Guard~\cite{fang_next-guard_2026}, HIDDENGUARD~\cite{mei_hiddenguard_2024}, and Kelp~\cite{li_kelp_2025} all move safety monitoring toward streaming or latent-state signals. Our work is aligned with that trajectory, but differs in two ways. First, we focus specifically on implicit harmful dialogue rather than broad harmfulness detection. Second, we impose a same-pass constraint: the monitor may read the generator's own hidden states during ordinary decoding, but may not invoke an additional forward pass through the base model. On the supervision side, FineHarm~\cite{li_judgment_2025} is useful because it provides onset-sensitive labels over explicit harmful continuation rather than only response-level labels, while Anthropic HH-RLHF~\cite{bai_training_2022} and Anthropic red-teaming data~\cite{ganguli_red_2022} provide matched safe and unsafe behavior that we use for support-style supervision and prompt-conditioned residual learning.

\section{Method}

\begin{figure}[t]
    \centering
    \includegraphics[width=0.85\linewidth]{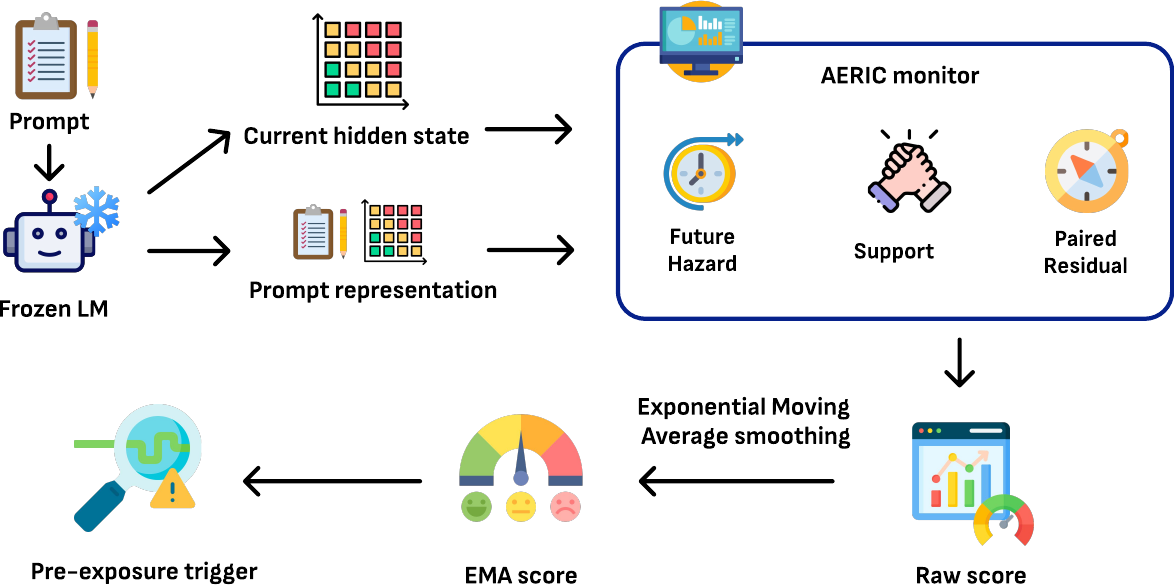}
    \caption{
    Overview of \textsc{AERIC}. During ordinary decoding, the frozen generator produces a current hidden state $h_t$ and a cached prompt representation $p$. \textsc{AERIC} reads these already-computed states, computes future-hazard, support, and paired-residual scores, and applies EMA smoothing to produce an online trigger signal.
    }\label{fig:aeric_overview}
\end{figure}

\subsection{Same-Pass Monitoring Setup}

Let a frozen causal language model generate tokens $y_1,\ldots,y_T$ conditioned on a prompt $x$. At decode step $t$, the model produces a hidden state $h_t \in \mathbb{R}^d$ for the current prefix $(x,y_{<t})$. We also cache a prompt summary $p \in \mathbb{R}^d$ from prompt-side hidden states before response decoding begins. A monitor is same-pass if its decision at time $t$ depends only on quantities already available from this decode trajectory, such as $h_t$, $p$, and previous monitor values. It may not invoke an additional forward pass through the base model or repeatedly call a separate generative guard. This pipeline is depicted in Figure~\ref{fig:aeric_overview}.

This constraint defines the deployment setting we care about. The monitor must be prefix-measurable, available before end-of-sequence, and cheap enough to run during ordinary decoding. In the default \textsc{AERIC} monitor, hidden states are projected to a $128$-dimensional representation and standardized. The trainable part of the monitor consists only of three scalar heads, described below.

\subsection{Anticipatory Evidence and Prompt-Conditioned Drift}

\textsc{AERIC} combines three signals. The first signal is future hazard. Instead of training a standard unsafe-prefix classifier that labels every prefix of an unsafe response as unsafe, we use onset-aware supervision from FineHarm~\cite{li_judgment_2025}. Let $o$ denote the harmful onset index derived from span or sentence annotations. For a horizon $H$, the hazard label at prefix $t$ is
\begin{equation}
    z_t^{\mathrm{haz}} = \mathbb{1}[\,0 \le o - t \le H\,].
\end{equation}
The future-hazard score is
\begin{equation}
    f_t = w_f^\top \tilde{h}_t + b_f,
\end{equation}
where $\tilde{h}_t$ is the projected and standardized hidden state. This head asks whether the current trajectory is close to entering harmful continuation, rather than whether the visible prefix already looks unsafe.

The second signal is support evidence. A hazard-only monitor can overfire on emotionally intense but still bounded responses, especially in advice, refusal, or de-escalation settings. We therefore train a support head
\begin{equation}
    c_t = w_c^\top \tilde{h}_t + b_c
\end{equation}
whose positive class consists of safe, supportive, bounded, or de-escalatory continuations, and whose negative class consists of imminent or post-onset unsafe continuations. In practice, the support-positive pool mixes safe auxiliary rows with harmless responses to harmful prompts from Anthropic HH-style data~\cite{bai_training_2022}. This head acts as counterevidence: it suppresses alerts when the response is intense but still consistent with safe assistance.

The third signal is prompt-conditioned residual drift. A simple way to condition on the prompt would be to classify $h_t-p$ directly, but naive prompt subtraction can remove useful structure. Instead, we learn the residual direction from matched safe and unsafe continuations for the same prompt. Let $h_t^{(s)}$ and $h_t^{(u)}$ denote hidden states from safe and unsafe continuations paired at the prompt level, and define
\begin{equation}
    \delta_t^{(s)} = h_t^{(s)} - p,
    \qquad
    \delta_t^{(u)} = h_t^{(u)} - p.
\end{equation}
The residual head scores a prompt-relative state as
\begin{equation}
    r(\delta) = w_r^\top \delta + b_r,
\end{equation}
and is trained with the pairwise ranking loss
\begin{equation}
    \mathcal{L}_{\mathrm{res}} =
    \max\left(0,\ 1 - r(\delta_t^{(u)}) + r(\delta_t^{(s)})\right).
\end{equation}
This encourages unsafe residual states to score above safe residual states for the same prompt. At test time, the residual score is
\begin{equation}
    r_t = r(h_t - p).
\end{equation}
The residual head is therefore not just prompt subtraction. It learns which prompt-relative deviations correspond to unsafe continuation under matched prompt conditions.

\subsection{Online Score, Size, and Calibration}

The raw per-token monitor score combines the three signals:
\begin{equation}
    g_t = f_t - \alpha c_t + \beta r_t,
\end{equation}
where $\alpha>0$ controls suppression by support evidence and $\beta>0$ controls the contribution of prompt-conditioned residual drift. The future-hazard term asks whether harmful continuation is likely soon, the support term asks whether the current trajectory still looks bounded or corrective, and the residual term asks whether the continuation has drifted toward an unsafe direction relative to safe behavior for the same prompt type.

For online monitoring, we smooth the raw score with an exponential moving average:
\begin{equation}
    m_t = \lambda g_t + (1-\lambda)m_{t-1},
\end{equation}
with $\lambda=0.3$ in our main runs. Appendix~\ref{app:monitor_hyperparams} includes a short sensitivity check showing that balanced AUROC is stable across a wider range of $\lambda$ values. A generation is flagged when
\begin{equation}
    m_t \ge \tau.
\end{equation}
Because $m_t$ depends only on the current hidden state, the cached prompt summary, and previous monitor values, the decision remains same-pass and prefix-measurable.

The default linear monitor is intentionally small. Each of the three scalar heads has $128$ weights and one bias, for a total of $3 \times 129 = 387$ trainable head parameters. The deployed artifact additionally stores fixed projection matrices and normalization statistics, totaling about $1.60$M stored scalars ($6.1$ MiB in fp32) for \texttt{Qwen/Qwen3{-}8B} and about $1.00$M stored scalars ($3.8$ MiB) for \texttt{google/gemma{-}4{-}E4B{-}it}. These counts describe only the monitor artifact; the generator is the frozen model whose already-computed hidden states are being read. Full monitor-size and hyperparameter details are reported in Appendix~\ref{app:monitor_hyperparams}.

For balanced benchmarks, no operating threshold is selected; AUROC and AUPRC are computed from the terminal EMA score. For prompt-level trigger benchmarks, we use a source-side safe-budget calibration rule for \textsc{AERIC}: on a held-out source calibration split, we choose the threshold that maximizes trigger coverage while constraining the safe-trigger rate to be at most $10\%$. We therefore treat prompt-benchmark results as exposure-withholding diagnostics rather than as primary balanced-classification evidence. We also tested small same-pass MLP replacements for individual heads, but report them only as a head-capacity follow-up because they were not standardized across all model and benchmark pairs.

\section{Experimental Setup}

\subsection{Model and Inference}

We evaluate \textsc{AERIC} on two base models: \texttt{Qwen3{-}8B} and \texttt{gemma{-}4{-}E4B{-}it}. In both cases we read hidden states from layer $-8$. The monitor is always same-pass: no experiment allows an additional forward pass through the generator. The default \textsc{AERIC} scores come from small linear heads on top of hidden states that are already available during decoding, and the online decision signal is the EMA-smoothed monitor described above.

\subsection{Source and Target Data}

\paragraph{Source supervision.}
Most transfer experiments use a FineHarm split of $1024 / 256 / 256$ for train, calibration, and held-out source test. The paired residual head uses $1024$ matched safe rows and $1024$ matched unsafe rows from Anthropic HH data, yielding $981$ matched prompt contexts. We also use auxiliary safe rows from English advice-oriented sources when training the support head.

\paragraph{Balanced implicit-harm targets.}
We evaluate on the full DiaSafety\cite{sun_safety_2022} test set ($1094$ rows) and on Harmful Advice\cite{luettgau2025harmfuladvice, luettgau_people_2026} ($550$ rows). DiaSafety is the harder conversational target because contextual confounds are stronger. Harmful Advice is the clearest transfer win for the full method.

\paragraph{Prompt-conditioned trigger targets.}
We evaluate on HarmBench DirectRequest~\cite{mazeika_harmbench_2024} ($320$ prompts) and on SocialHarmBench~\cite{pandey_socialharmbench_2026} ($585$ raw prompts; $584$ non-empty prompts in our run). Both are prompt-only harmful request suites, so they are suitable for answer-generation trigger diagnostics but not for balanced response classification. Their role here is purely systems-oriented: does the monitor fire early enough during generation to withhold continuation before it is exposed. Additional dataset and preprocessing details are given in Appendix~\ref{app:dataset_preprocessing}.

\subsection{Baselines}

We compare against three external baselines.
\begin{itemize}[leftmargin=1.5em]
    \item \textbf{Post-hoc moderation}: ShieldGemma{-}9B and WildGuard~\cite{han_wildguard_2024} on completed prompt-response pairs. For prompt-level trigger comparisons, we run ShieldGemma and WildGuard on growing prefixes as fixed-budget streaming approximations.
    \item \textbf{Native streaming guard}: Qwen3Guard{-}Stream{-}4B~\cite{zhao_qwen3guard_2025}, a pretrained token-level streaming guard evaluated on both balanced targets, on HarmBench DirectRequest, and on SocialHarmBench for both generator families.
\end{itemize}

\subsection{Metrics}

\paragraph{Balanced implicit-harm benchmarks.}
On DiaSafety and Harmful Advice we report AUROC and AUPRC in the main table. These are the paper's primary threshold-free ranking metrics. Each balanced-benchmark cell additionally reports a nonparametric $95\%$ bootstrap confidence interval obtained by resampling evaluation rows with replacement and recomputing the metric. The intervals therefore capture evaluation-set sampling variability with the trained monitor and frozen generator held fixed, rather than retraining or random-seed variation. Thresholded operating behavior still matters for deployment, but it is secondary to the ranking story and is discussed in the text rather than elevated to the headline table.

\paragraph{Prompt-conditioned trigger benchmarks.}
On HarmBench DirectRequest and SocialHarmBench we report trigger@$8$, trigger@$16$, trigger@$32$, trigger@$64$, and mean \emph{withheld answer tokens} relative to a no-stop baseline. Here, withheld answer tokens refers to the number of answer tokens that would have been emitted under no-stop but were never shown because the monitor fired first. For \textsc{AERIC}, we set the operating threshold by a source-side safe-budget calibration rule: on a held-out source calibration split, we choose the threshold that maximizes trigger coverage while constraining the safe-trigger rate to be at most $10\%$. The calibration objective and selected source-side thresholds are detailed in Appendix~\ref{app:calibration_details}.

\section{Results}

\subsection{Balanced Implicit-Harm Benchmarks}

\begin{table*}[t]
\centering
\small
\setlength{\tabcolsep}{4pt}
\renewcommand{\arraystretch}{1.15}
\begin{tabular}{lcccc}
\toprule
\multirow{2}{*}{Method} & \multicolumn{2}{c}{DiaSafety} & \multicolumn{2}{c}{Harmful Advice} \\
\cmidrule(lr){2-3} \cmidrule(lr){4-5}
 & AUROC & AUPRC & AUROC & AUPRC \\
\midrule
\multicolumn{5}{l}{\textbf{\textsc{AERIC}}} \\
\textsc{AERIC} (\texttt{Qwen3{-}8B}) & \metricci{\underline{0.7143}}{[0.6838, 0.7430]} & \metricci{0.6341}{[0.5988, 0.6730]} & \metricci{\textbf{0.8582}}{[0.8217, 0.8928]} & \metricci{\textbf{0.9605}}{[0.9483, 0.9715]} \\
\textsc{AERIC} (\texttt{Gemma{-}4{-}E4B{-}it}) & \metricci{\textbf{0.7181}}{[0.6882, 0.7468]} & \metricci{\textbf{0.6632}}{[0.6277, 0.6993]} & \metricci{\underline{0.8287}}{[0.7855, 0.8689]} & \metricci{0.9442}{[0.9234, 0.9624]} \\
\addlinespace[2pt]
\multicolumn{5}{l}{\textbf{External baselines}} \\
Qwen3Guard{-}Stream{-}4B & \metricci{0.6830}{[0.6514, 0.7132]} & \metricci{\underline{0.6411}}{[0.6053, 0.6808]} & \metricci{0.8219}{[0.7827, 0.8581]} & \metricci{\underline{0.9515}}{[0.9395, 0.9624]} \\
WildGuard-7B & \metricci{0.6855}{[0.6585, 0.7121]} & \metricci{0.6009}{[0.5743, 0.6289]} & \metricci{0.6784}{[0.6568, 0.7011]} & \metricci{0.8714}{[0.8627, 0.8805]} \\
ShieldGemma{-}9B & \metricci{0.4901}{[0.4556, 0.5240]} & \metricci{0.4731}{[0.4444, 0.5050]} & \metricci{0.3709}{[0.3174, 0.4256]} & \metricci{0.7682}{[0.7434, 0.7941]} \\
\bottomrule
\end{tabular}
\renewcommand{\arraystretch}{1.0}
\caption{Balanced implicit-harm benchmarks. Each cell reports the point estimate with a nonparametric $95\%$ bootstrap confidence interval on a second line, computed by resampling evaluation rows with replacement. Bold marks the best point estimate in each metric column and underline marks the second-best point estimate. \textsc{AERIC} is evaluated with hidden states from two base models.}\label{tab:balanced_results}
\end{table*}

Table~\ref{tab:balanced_results} shows that \textsc{AERIC} improves implicit-harm ranking over both response-level guards and the strongest native streaming guard in our comparison. The main table reports point estimates together with nonparametric $95\%$ bootstrap confidence intervals for the balanced benchmarks. Compared with the best external baseline in each metric column, the best \textsc{AERIC} configuration improves on DiaSafety by $3.26$ AUROC points and $2.21$ AUPRC points, and on Harmful Advice by $3.63$ AUROC points and $0.90$ AUPRC points. The gains are largest in AUROC, indicating that the main effect is improved ordering of safe and unsafe continuations rather than simply choosing a better operating threshold. This distinction matters because our balanced benchmarks are evaluated as threshold-free ranking tasks, while deployment-time threshold selection is handled separately in the trigger experiments.

These results support the paper's main balanced-benchmark claim: explicit harmful supervision can be converted into a same-pass hidden-state signal that transfers to implicit harmful dialogue. The cross-model pattern strengthens this interpretation. Gemma gives the strongest DiaSafety result, while Qwen gives the strongest Harmful Advice result, so the monitor is not simply exploiting idiosyncrasies of one generator's representation space. Instead, the same future-support-residual monitor family extracts a reusable anticipatory risk signal from already-computed hidden states, improving over guards that rely on completed text or separate streaming guard computation.

\subsection{Early-Trigger Benchmarks}

Table~\ref{tab:prompt_trigger_results} reports prompt-only harmful request benchmarks separately because they are not balanced response-classification tasks. Their role is systems-oriented: can a source-calibrated online monitor fire early enough to withhold continuation before more of the answer is exposed?

\begin{table*}[t]
\centering
\small
\setlength{\tabcolsep}{4pt}
\begin{tabular}{llccccc}
\toprule
\multirow{2}{*}{Setting} & \multirow{2}{*}{Method} & \multicolumn{4}{c}{Trigger coverage} & \multirow{2}{*}{Mean Withheld} \\
\cmidrule(lr){3-6}
 &  & @8 & @16 & @32 & @64 & \\
\midrule
\multicolumn{7}{l}{\textbf{\textit{HarmBench DirectRequest}}} \\
\texttt{Qwen3{-}8B} & \textsc{AERIC} & \textbf{0.5750} & \textbf{0.5875} & \textbf{0.6125} & \textbf{0.6438} & \textbf{32.5469} \\
 & Qwen3Guard{-}Stream{-}4B & 0.0469 & 0.0531 & 0.0813 & 0.1063 & 4.3094 \\
 & WildGuard-7B Prefix & 0.1313 & 0.1438 & 0.1438 & 0.1438 & 7.3125 \\
 & ShieldGemma{-}9B Prefix & 0.0312 & 0.0781 & 0.1656 & 0.2188 & 6.7625 \\
\addlinespace[1pt]
\texttt{Gemma{-}4{-}E4B{-}it} & \textsc{AERIC} & \textbf{0.3438} & \textbf{0.3594} & \textbf{0.4062} & \textbf{0.4656} & \textbf{23.5281} \\
 & Qwen3Guard{-}Stream{-}4B & 0.0438 & 0.0750 & 0.1531 & 0.3281 & 9.3375 \\
 & WildGuard-7B Prefix & 0.2219 & 0.2531 & 0.2875 & 0.3750 & 15.3313 \\
 & ShieldGemma{-}9B Prefix & 0.0281 & 0.0688 & 0.1469 & 0.2094 & 6.3937 \\
\addlinespace[2pt]
\multicolumn{7}{l}{\textbf{\textit{SocialHarmBench}}} \\
\texttt{Qwen3{-}8B} & \textsc{AERIC} & \textbf{0.6353} & \textbf{0.6438} & \textbf{0.6592} & \textbf{0.6849} & \textbf{32.7688} \\
 & Qwen3Guard{-}Stream{-}4B & 0.0274 & 0.0394 & 0.0651 & 0.0719 & 3.4110 \\
 & WildGuard Prefix & 0.1130 & 0.1182 & 0.1199 & 0.1199 & 6.5565 \\
 & ShieldGemma{-}9B Prefix & 0.0120 & 0.0325 & 0.1147 & 0.2038 & 5.3904 \\
\addlinespace[1pt]
\texttt{Gemma{-}4{-}E4B{-}it} & \textsc{AERIC} & \textbf{0.6764} & \textbf{0.6781} & \textbf{0.6918} & \textbf{0.7363} & \textbf{41.8579} \\
 & Qwen3Guard{-}Stream{-}4B & 0.0599 & 0.1079 & 0.1764 & 0.2962 & 10.0856 \\
 & WildGuard Prefix & 0.2671 & 0.2723 & 0.2723 & 0.3322 & 15.2911 \\
 & ShieldGemma{-}9B Prefix & 0.0120 & 0.0171 & 0.0668 & 0.1199 & 2.7432 \\
\bottomrule
\end{tabular}
\caption{Prompt-level trigger benchmarks under a source-side $10\%$ safe-trigger budget for \textsc{AERIC}. Rows are grouped first by dataset and then by evaluation setting; bold marks the best value within each dataset/setting block. Withheld denotes the mean number of answer tokens that would have been shown under no-stop but were never exposed because the monitor fired first. The safe-budget calibration applies to \textsc{AERIC}; external guards use their native unsafe decisions under the prefix protocol described in Appendix~\ref{app:baseline_protocols}}\label{tab:prompt_trigger_results}.
\end{table*}

The prompt-level picture is more conservative under the source-side $10\%$ safe-trigger budget, but it remains encouraging. On HarmBench DirectRequest with \texttt{Qwen3{-}8B}, \textsc{AERIC} reaches trigger@$64$ $0.6438$, compared with $0.1063$ for Qwen3Guard{-}Stream{-}4B, $0.1438$ for WildGuard Prefix, and $0.2188$ for ShieldGemma{-}9B Prefix, while withholding $32.55$ answer tokens on average. On the same HarmBench slice with \texttt{Gemma{-}4{-}E4B{-}it}, \textsc{AERIC} reaches $0.4656$, compared with $0.3281$ for Qwen3Guard{-}Stream{-}4B, $0.3750$ for WildGuard Prefix, and $0.2094$ for ShieldGemma{-}9B Prefix. SocialHarmBench shows that the signal is not limited to generic harmful instructions and that the effect survives a second generator family. Under the Qwen generator, the same calibrated monitor reaches trigger@$64$ $0.6849$, compared with $0.0719$ for Qwen3Guard{-}Stream{-}4B, $0.1199$ for WildGuard Prefix, and $0.2038$ for ShieldGemma{-}9B Prefix. Under the Gemma generator, it reaches $0.7363$, compared with $0.2962$ for Qwen3Guard{-}Stream{-}4B, $0.3322$ for WildGuard Prefix, and $0.1199$ for ShieldGemma{-}9B Prefix, while withholding $41.86$ answer tokens on average. These benchmarks depend on the source-side $10\%$ safe-trigger budget calibration and therefore should not be read as balanced classification evidence, but they do show that the same-pass signal remains actionable before end-of-sequence on prompt-only harmful request suites.

\subsection{Component Ablation of the Base Score Family}

The final online monitor combines the three linear heads with an EMA decision rule. To isolate what each head contributes under that same online rule, Table~\ref{tab:ablation} reruns the underlying \texttt{Qwen/Qwen3{-}8B} transfer family one component at a time with the refreshed EMA aggregation: future-hazard only, then future-hazard plus support, and finally the paired residual head. The main balanced results above are the final EMA-selected monitor; this table is the matching structural ablation under the same aggregation scheme.

\begin{table}[t]
\centering
\small
\setlength{\tabcolsep}{5pt}
\begin{tabular}{lcccc}
\toprule
\multirow{2}{*}{Method} & \multicolumn{2}{c}{DiaSafety} & \multicolumn{2}{c}{Harmful Advice} \\
\cmidrule(lr){2-3} \cmidrule(lr){4-5}
 & AUROC & AUPRC & AUROC & AUPRC \\
\midrule
future only & 0.6839 & 0.6150 & 0.6873 & 0.8829 \\
future + support & 0.6993 & 0.6222 & 0.8430 & 0.9559 \\
full monitor (future + support + residual) & \textbf{0.7143} & \textbf{0.6341} & \textbf{0.8582} & \textbf{0.9605} \\
\bottomrule
\end{tabular}
\caption{Ablation of the underlying \texttt{Qwen/Qwen3{-}8B} transfer family under the same EMA aggregation used by the final online monitor. Bold marks the best value in each metric column. The support head provides the first large gain on Harmful Advice, and the paired residual term adds the final AUROC lift on both balanced targets.}\label{tab:ablation}
\end{table}

The pattern is consistent across targets. The support head explains the first large jump on Harmful Advice, where assertive but still bounded guidance is a major confound, moving AUROC from $0.6873$ to $0.8430$. The paired residual term then improves AUROC on both targets, with the clearest relative benefit on DiaSafety, where prompt-conditioned conversational confounds are stronger, moving AUROC from $0.6993$ to $0.7143$. DiaSafety AUPRC also rises modestly after adding the residual head, which suggests that the main benefit is better ranking without changing the basic mechanism of the monitor family. In other words, the final EMA monitor is not hiding a different signal; it is aggregating a score family whose component contributions still point in the same direction under the refreshed methodology.

\subsection{Computational Overhead}

The systems story is equally important. We measure latency on a harmful-prompt fixed-generation overhead benchmark sampled from HarmBench DirectRequest and SocialHarmBench under \texttt{Qwen/Qwen3{-}8B}. On that aggregate, the paired-residual monitor increases mean latency by only $2.34\%$, compared with $79.40\%$ for Qwen3Guard{-}Stream{-}4B, $158.73\%$ for prefixized ShieldGemma{-}9B, and $216.74\%$ for prefixized WildGuard. Broken out by dataset, \textsc{AERIC} increases mean latency by $2.38\%$ on HarmBench DirectRequest and $2.31\%$ on SocialHarmBench; the corresponding numbers are $77.61\%$ and $81.41\%$ for Qwen3Guard{-}Stream{-}4B, $157.68\%$ and $159.91\%$ for prefixized ShieldGemma{-}9B, and $218.88\%$ and $214.35\%$ for prefixized WildGuard. Measurement details are reported in Appendix~\ref{app:systems_measurement}.

\section{Limitations and Broader Impact}

\textsc{AERIC} should be understood as an online risk signal, not as a complete intervention policy. Our balanced results evaluate threshold-free ranking, while prompt-level trigger results on HarmBench DirectRequest and SocialHarmBench depend on a frozen operating threshold selected before evaluation. High trigger coverage can reduce exposure, but it does not decide what a deployed system should do next: stop generation, ask a clarifying question, route to a safer response policy, or escalate to human review. This distinction matters because safety interventions can create their own failure modes. In particular, overly conservative safeguards can produce false refusals or over-refusal on benign but sensitive requests, reducing usefulness precisely in settings where users may need careful support~\cite{cui_or-bench_2025,zhang_falsereject_2025,sullutrone_cover_2025}. The support head is designed to reduce part of this failure mode, but it does not remove the need for careful thresholding and downstream policy design.

The empirical scope is also limited. We evaluate two generator families and read hidden states from a single layer, so the current results do not establish robustness across model scales, architectures, languages, decoding strategies, or layer choices. The supervision is also imperfectly matched to the target setting: we transfer from explicit harmful supervision and matched safe/unsafe behavior to implicit, context-sensitive harmfulness, rather than training on large-scale prompt-matched implicit-harm onset labels. Better source data with implicit onset annotations, more domain-matched safe continuations, and multilingual coverage would likely improve the transfer frontier. More broadly, same-pass monitoring reduces overhead but does not remove safety risk: a low-latency hidden-state monitor can warn before further exposure, but automatic hard stops can still produce confusing truncations, unnecessary refusals, or missed harms. \textsc{AERIC} should therefore be treated as one component of a broader safety stack rather than as a standalone substitute for evaluation, user-facing policy design, and human oversight~\cite{noauthor_international_nodate}.

\section{Conclusion}

We presented \textsc{AERIC}, a same-pass hidden-state monitor for implicit harmful dialogue. Rather than asking a separate guard to repeatedly classify growing text prefixes, \textsc{AERIC} reads the generator's already-computed hidden states and combines short-horizon hazard forecasting, support-sensitive suppression, and prompt-conditioned residual scoring under an EMA-smoothed online decision rule. The results show that explicit harmful supervision can transfer to implicit-harm ranking across two generator families, improving over response-level guards and Qwen3Guard{-}Stream{-}4B, the strongest native streaming baseline in our comparison. The same frozen monitor also remains actionable on prompt-only harmful request suites, where it can trigger before end-of-sequence and withhold continuation before further exposure. The claim is deliberately limited: \textsc{AERIC} provides a lightweight pre-exposure risk signal, not a complete serving policy. Remaining bottlenecks include contextual confounds on DiaSafety, calibration-sensitive trigger behavior, and the still-open problem of deciding how an application should respond after an early risk signal.

\bibliographystyle{unsrtnat}
\bibliography{references}

@misc{li_judgment_2025,
	title = {From {Judgment} to {Interference}: {Early} {Stopping} {LLM} {Harmful} {Outputs} via {Streaming} {Content} {Monitoring}},
	shorttitle = {From {Judgment} to {Interference}},
	url = {http://arxiv.org/abs/2506.09996},
	doi = {10.48550/arXiv.2506.09996},
	urldate = {2026-04-22},
	publisher = {arXiv},
	author = {Li, Yang and Sheng, Qiang and Yang, Yehan and Zhang, Xueyao and Cao, Juan},
	month = sep,
	year = {2025},
	note = {arXiv:2506.09996 [cs]},
}

@misc{kavumba_predict_2026,
	title = {Predict, {Don}'t {React}: {Value}-{Based} {Safety} {Forecasting} for {LLM} {Streaming}},
	shorttitle = {Predict, {Don}'t {React}},
	url = {http://arxiv.org/abs/2604.03962},
	doi = {10.48550/arXiv.2604.03962},
	urldate = {2026-04-22},
	publisher = {arXiv},
	author = {Kavumba, Pride and Wataoka, Koki and Nguyen, Huy H. and Li, Jiaxuan and Ohagi, Masaya},
	month = apr,
	year = {2026},
	note = {arXiv:2604.03962 [cs]},
}

@inproceedings{xuan_shieldhead_2025,
	address = {Vienna, Austria},
	title = {{ShieldHead}: {Decoding}-time {Safeguard} for {Large} {Language} {Models}},
	isbn = {979-8-89176-256-5},
	shorttitle = {{ShieldHead}},
	url = {https://aclanthology.org/2025.findings-acl.932/},
	doi = {10.18653/v1/2025.findings-acl.932},
	urldate = {2026-04-22},
	booktitle = {Findings of the {Association} for {Computational} {Linguistics}: {ACL} 2025},
	publisher = {Association for Computational Linguistics},
	author = {Xuan, Zitao and Mao, Xiaofeng and Chen, Da and Zhang, Xin and Dong, Yuhan and Zhou, Jun},
	editor = {Che, Wanxiang and Nabende, Joyce and Shutova, Ekaterina and Pilehvar, Mohammad Taher},
	month = jul,
	year = {2025},
	pages = {18129--18143},
}

@inproceedings{mei_mitigating_2022,
	address = {Abu Dhabi, United Arab Emirates},
	title = {Mitigating {Covertly} {Unsafe} {Text} within {Natural} {Language} {Systems}},
	url = {https://aclanthology.org/2022.findings-emnlp.211/},
	doi = {10.18653/v1/2022.findings-emnlp.211},
	urldate = {2026-04-22},
	booktitle = {Findings of the {Association} for {Computational} {Linguistics}: {EMNLP} 2022},
	publisher = {Association for Computational Linguistics},
	author = {Mei, Alex and Kabir, Anisha and Levy, Sharon and Subbiah, Melanie and Allaway, Emily and Judge, John and Patton, Desmond and Bimber, Bruce and McKeown, Kathleen and Wang, William Yang},
	editor = {Goldberg, Yoav and Kozareva, Zornitsa and Zhang, Yue},
	month = dec,
	year = {2022},
	pages = {2914--2926},
}

@inproceedings{sun_safety_2022,
	address = {Dublin, Ireland},
	title = {On the {Safety} of {Conversational} {Models}: {Taxonomy}, {Dataset}, and {Benchmark}},
	shorttitle = {On the {Safety} of {Conversational} {Models}},
	url = {https://aclanthology.org/2022.findings-acl.308/},
	doi = {10.18653/v1/2022.findings-acl.308},
	urldate = {2026-04-22},
	booktitle = {Findings of the {Association} for {Computational} {Linguistics}: {ACL} 2022},
	publisher = {Association for Computational Linguistics},
	author = {Sun, Hao and Xu, Guangxuan and Deng, Jiawen and Cheng, Jiale and Zheng, Chujie and Zhou, Hao and Peng, Nanyun and Zhu, Xiaoyan and Huang, Minlie},
	editor = {Muresan, Smaranda and Nakov, Preslav and Villavicencio, Aline},
	month = may,
	year = {2022},
	pages = {3906--3923},
}

@misc{qiu_benchmark_2023,
	title = {A {Benchmark} for {Understanding} {Dialogue} {Safety} in {Mental} {Health} {Support}},
	url = {http://arxiv.org/abs/2307.16457},
	doi = {10.48550/arXiv.2307.16457},
	urldate = {2026-04-22},
	publisher = {arXiv},
	author = {Qiu, Huachuan and Zhao, Tong and Li, Anqi and Zhang, Shuai and He, Hongliang and Lan, Zhenzhong},
	month = jul,
	year = {2023},
	note = {arXiv:2307.16457 [cs]},
}

@misc{wen_unveiling_2023,
	title = {Unveiling the {Implicit} {Toxicity} in {Large} {Language} {Models}},
	url = {http://arxiv.org/abs/2311.17391},
	doi = {10.48550/arXiv.2311.17391},
	urldate = {2026-04-22},
	publisher = {arXiv},
	author = {Wen, Jiaxin and Ke, Pei and Sun, Hao and Zhang, Zhexin and Li, Chengfei and Bai, Jinfeng and Huang, Minlie},
	month = nov,
	year = {2023},
	note = {arXiv:2311.17391 [cs]},
}

@inproceedings{zeng_root_2025,
	address = {Vienna, Austria},
	title = {Root {Defense} {Strategies}: {Ensuring} {Safety} of {LLM} at the {Decoding} {Level}},
	isbn = {979-8-89176-251-0},
	shorttitle = {Root {Defense} {Strategies}},
	url = {https://aclanthology.org/2025.acl-long.97/},
	doi = {10.18653/v1/2025.acl-long.97},
	urldate = {2026-04-23},
	booktitle = {Proceedings of the 63rd {Annual} {Meeting} of the {Association} for {Computational} {Linguistics} ({Volume} 1: {Long} {Papers})},
	publisher = {Association for Computational Linguistics},
	author = {Zeng, Xinyi and Shang, Yuying and Chen, Jiawei and Zhang, Jingyuan and Tian, Yu},
	editor = {Che, Wanxiang and Nabende, Joyce and Shutova, Ekaterina and Pilehvar, Mohammad Taher},
	month = jul,
	year = {2025},
	pages = {1974--1988},
}

@inproceedings{phan_think_2025,
	address = {Suzhou, China},
	title = {Think {Twice}, {Generate} {Once}: {Safeguarding} by {Progressive} {Self}-{Reflection}},
	isbn = {979-8-89176-335-7},
	shorttitle = {Think {Twice}, {Generate} {Once}},
	url = {https://aclanthology.org/2025.findings-emnlp.503/},
	doi = {10.18653/v1/2025.findings-emnlp.503},
	urldate = {2026-04-23},
	booktitle = {Findings of the {Association} for {Computational} {Linguistics}: {EMNLP} 2025},
	publisher = {Association for Computational Linguistics},
	author = {Phan, Hoang and Li, Victor and Lei, Qi},
	editor = {Christodoulopoulos, Christos and Chakraborty, Tanmoy and Rose, Carolyn and Peng, Violet},
	month = nov,
	year = {2025},
	pages = {9466--9483},
}

@misc{jiao_llm_2026,
	title = {{LLM} {Safety} {From} {Within}: {Detecting} {Harmful} {Content} with {Internal} {Representations}},
	shorttitle = {{LLM} {Safety} {From} {Within}},
	url = {http://arxiv.org/abs/2604.18519},
	doi = {10.48550/arXiv.2604.18519},
	urldate = {2026-04-23},
	publisher = {arXiv},
	author = {Jiao, Difan and Liu, Yilun and Yuan, Ye and Tang, Zhenwei and Du, Linfeng and Wu, Haolun and Anderson, Ashton},
	month = apr,
	year = {2026},
	note = {arXiv:2604.18519 [cs]
version: 1},
}

@misc{yang_prefix_2025,
	title = {Prefix {Probing}: {Lightweight} {Harmful} {Content} {Detection} for {Large} {Language} {Models}},
	shorttitle = {Prefix {Probing}},
	url = {http://arxiv.org/abs/2512.16650},
	doi = {10.48550/arXiv.2512.16650},
	urldate = {2026-04-23},
	publisher = {arXiv},
	author = {Yang, Jirui and Guo, Hengqi and Lu, Zhihui and Zhao, Yi and Zhang, Yuansen and Hu, Shijing and Duan, Qiang and Wang, Yinggui and Wei, Tao},
	month = dec,
	year = {2025},
	note = {arXiv:2512.16650 [cs]
version: 1},
}

@dataset{luettgau2025harmfuladvice,
  title={Harmful Advice Dataset},
  author={Luettgau, Lennart and Davidson, Henry and Nguyen, Elizabeth and Butuc, Daria and Summerfield, Christopher},
  year={2025},
  institution={UK AI Security Institute},
  url={https://huggingface.co/datasets/ai-safety-institute/harmful-advice-dataset}
}

@inproceedings{pal_future_2023,
	address = {Singapore},
	title = {Future {Lens}: {Anticipating} {Subsequent} {Tokens} from a {Single} {Hidden} {State}},
	shorttitle = {Future {Lens}},
	url = {https://aclanthology.org/2023.conll-1.37/},
	doi = {10.18653/v1/2023.conll-1.37},
	urldate = {2026-04-23},
	booktitle = {Proceedings of the 27th {Conference} on {Computational} {Natural} {Language} {Learning} ({CoNLL})},
	publisher = {Association for Computational Linguistics},
	author = {Pal, Koyena and Sun, Jiuding and Yuan, Andrew and Wallace, Byron and Bau, David},
	editor = {Jiang, Jing and Reitter, David and Deng, Shumin},
	month = dec,
	year = {2023},
	pages = {548--560},
}

@inproceedings{azaria_internal_2023,
	address = {Singapore},
	title = {The {Internal} {State} of an {LLM} {Knows} {When} {It}'s {Lying}},
	url = {https://aclanthology.org/2023.findings-emnlp.68/},
	doi = {10.18653/v1/2023.findings-emnlp.68},
	urldate = {2026-04-23},
	booktitle = {Findings of the {Association} for {Computational} {Linguistics}: {EMNLP} 2023},
	publisher = {Association for Computational Linguistics},
	author = {Azaria, Amos and Mitchell, Tom},
	editor = {Bouamor, Houda and Pino, Juan and Bali, Kalika},
	month = dec,
	year = {2023},
	pages = {967--976},
}

@misc{han_wildguard_2024,
	title = {{WildGuard}: {Open} {One}-{Stop} {Moderation} {Tools} for {Safety} {Risks}, {Jailbreaks}, and {Refusals} of {LLMs}},
	shorttitle = {{WildGuard}},
	url = {http://arxiv.org/abs/2406.18495},
	doi = {10.48550/arXiv.2406.18495},
	urldate = {2026-04-23},
	publisher = {arXiv},
	author = {Han, Seungju and Rao, Kavel and Ettinger, Allyson and Jiang, Liwei and Lin, Bill Yuchen and Lambert, Nathan and Choi, Yejin and Dziri, Nouha},
	month = dec,
	year = {2024},
	note = {arXiv:2406.18495 [cs]},
}

@misc{zhao_qwen3guard_2025,
	title = {{Qwen3Guard} {Technical} {Report}},
	url = {http://arxiv.org/abs/2510.14276},
	doi = {10.48550/arXiv.2510.14276},
	urldate = {2026-04-23},
	publisher = {arXiv},
	author = {Zhao, Haiquan and Yuan, Chenhan and Huang, Fei and Hu, Xiaomeng and Zhang, Yichang and Yang, An and Yu, Bowen and Liu, Dayiheng and Zhou, Jingren and Lin, Junyang and Yang, Baosong and Cheng, Chen and Tang, Jialong and Jiang, Jiandong and Zhang, Jianwei and Xu, Jijie and Yan, Ming and Sun, Minmin and Zhang, Pei and Xie, Pengjun and Tang, Qiaoyu and Zhu, Qin and Zhang, Rong and Wu, Shibin and Zhang, Shuo and He, Tao and Tang, Tianyi and Xia, Tingyu and Liao, Wei and Shen, Weizhou and Yin, Wenbiao and Zhou, Wenmeng and Yu, Wenyuan and Wang, Xiaobin and Deng, Xiaodong and Xu, Xiaodong and Zhang, Xinyu and Liu, Yang and Li, Yeqiu and Zhang, Yi and Jiang, Yong and Wan, Yu and Zhou, Yuxin},
	month = oct,
	year = {2025},
	note = {arXiv:2510.14276 [cs]},
}

@misc{fang_next-guard_2026,
	title = {{NExT}-{Guard}: {Training}-{Free} {Streaming} {Safeguard} without {Token}-{Level} {Labels}},
	shorttitle = {{NExT}-{Guard}},
	url = {http://arxiv.org/abs/2603.02219},
	doi = {10.48550/arXiv.2603.02219},
	urldate = {2026-04-23},
	publisher = {arXiv},
	author = {Fang, Junfeng and Chen, Nachuan and Jiang, Houcheng and Zhang, Dan and Shen, Fei and Wang, Xiang and He, Xiangnan and Chua, Tat-Seng},
	month = feb,
	year = {2026},
	note = {arXiv:2603.02219 [cs]},
}

@misc{mei_hiddenguard_2024,
	title = {{HiddenGuard}: {Fine}-{Grained} {Safe} {Generation} with {Specialized} {Representation} {Router}},
	shorttitle = {{HiddenGuard}},
	url = {http://arxiv.org/abs/2410.02684},
	doi = {10.48550/arXiv.2410.02684},
	urldate = {2026-04-23},
	publisher = {arXiv},
	author = {Mei, Lingrui and Liu, Shenghua and Wang, Yiwei and Bi, Baolong and Yuan, Ruibin and Cheng, Xueqi},
	month = oct,
	year = {2024},
	note = {arXiv:2410.02684 [cs]},
}

@misc{li_kelp_2025,
	title = {Kelp: {A} {Streaming} {Safeguard} for {Large} {Models} via {Latent} {Dynamics}-{Guided} {Risk} {Detection}},
	shorttitle = {Kelp},
	url = {http://arxiv.org/abs/2510.09694},
	doi = {10.48550/arXiv.2510.09694},
	urldate = {2026-04-23},
	publisher = {arXiv},
	author = {Li, Xiaodan and Wu, Mengjie and Zhu, Yao and Lv, Yunna and Chen, YueFeng and Chen, Cen and Guo, Jianmei and Xue, Hui},
	month = oct,
	year = {2025},
	note = {arXiv:2510.09694 [cs]},
}

@misc{bai_training_2022,
	title = {Training a {Helpful} and {Harmless} {Assistant} with {Reinforcement} {Learning} from {Human} {Feedback}},
	url = {http://arxiv.org/abs/2204.05862},
	doi = {10.48550/arXiv.2204.05862},
	urldate = {2026-04-23},
	publisher = {arXiv},
	author = {Bai, Yuntao and Jones, Andy and Ndousse, Kamal and Askell, Amanda and Chen, Anna and DasSarma, Nova and Drain, Dawn and Fort, Stanislav and Ganguli, Deep and Henighan, Tom and Joseph, Nicholas and Kadavath, Saurav and Kernion, Jackson and Conerly, Tom and El-Showk, Sheer and Elhage, Nelson and Hatfield-Dodds, Zac and Hernandez, Danny and Hume, Tristan and Johnston, Scott and Kravec, Shauna and Lovitt, Liane and Nanda, Neel and Olsson, Catherine and Amodei, Dario and Brown, Tom and Clark, Jack and McCandlish, Sam and Olah, Chris and Mann, Ben and Kaplan, Jared},
	month = apr,
	year = {2022},
	note = {arXiv:2204.05862 [cs]},
}

@misc{ganguli_red_2022,
	title = {Red {Teaming} {Language} {Models} to {Reduce} {Harms}: {Methods}, {Scaling} {Behaviors}, and {Lessons} {Learned}},
	shorttitle = {Red {Teaming} {Language} {Models} to {Reduce} {Harms}},
	url = {http://arxiv.org/abs/2209.07858},
	doi = {10.48550/arXiv.2209.07858},
	urldate = {2026-04-23},
	publisher = {arXiv},
	author = {Ganguli, Deep and Lovitt, Liane and Kernion, Jackson and Askell, Amanda and Bai, Yuntao and Kadavath, Saurav and Mann, Ben and Perez, Ethan and Schiefer, Nicholas and Ndousse, Kamal and Jones, Andy and Bowman, Sam and Chen, Anna and Conerly, Tom and DasSarma, Nova and Drain, Dawn and Elhage, Nelson and El-Showk, Sheer and Fort, Stanislav and Hatfield-Dodds, Zac and Henighan, Tom and Hernandez, Danny and Hume, Tristan and Jacobson, Josh and Johnston, Scott and Kravec, Shauna and Olsson, Catherine and Ringer, Sam and Tran-Johnson, Eli and Amodei, Dario and Brown, Tom and Joseph, Nicholas and McCandlish, Sam and Olah, Chris and Kaplan, Jared and Clark, Jack},
	month = nov,
	year = {2022},
	note = {arXiv:2209.07858 [cs]},
}

@misc{luettgau_people_2026,
	title = {People readily follow personal advice from {AI} but it does not improve their well-being},
	url = {http://arxiv.org/abs/2511.15352},
	doi = {10.48550/arXiv.2511.15352},
	urldate = {2026-04-23},
	publisher = {arXiv},
	author = {Luettgau, Lennart and Cheung, Vanessa and Dubois, Magda and Juechems, Keno and Bergs, Jessica and Symes, Luke and Davidson, Henry and O'Dell, Bessie and Kirk, Hannah Rose and Rollwage, Max and Summerfield, Christopher},
	month = apr,
	year = {2026},
	note = {arXiv:2511.15352 [cs]
version: 3},
}

@misc{hartvigsen_toxigen_2022,
	title = {{ToxiGen}: {A} {Large}-{Scale} {Machine}-{Generated} {Dataset} for {Adversarial} and {Implicit} {Hate} {Speech} {Detection}},
	shorttitle = {{ToxiGen}},
	url = {http://arxiv.org/abs/2203.09509},
	doi = {10.48550/arXiv.2203.09509},
	urldate = {2026-04-25},
	publisher = {arXiv},
	author = {Hartvigsen, Thomas and Gabriel, Saadia and Palangi, Hamid and Sap, Maarten and Ray, Dipankar and Kamar, Ece},
	month = jul,
	year = {2022},
	note = {arXiv:2203.09509 [cs]},
}

@misc{rauh_characteristics_2022,
	title = {Characteristics of {Harmful} {Text}: {Towards} {Rigorous} {Benchmarking} of {Language} {Models}},
	shorttitle = {Characteristics of {Harmful} {Text}},
	url = {http://arxiv.org/abs/2206.08325},
	doi = {10.48550/arXiv.2206.08325},
	urldate = {2026-04-25},
	publisher = {arXiv},
	author = {Rauh, Maribeth and Mellor, John and Uesato, Jonathan and Huang, Po-Sen and Welbl, Johannes and Weidinger, Laura and Dathathri, Sumanth and Glaese, Amelia and Irving, Geoffrey and Gabriel, Iason and Isaac, William and Hendricks, Lisa Anne},
	month = oct,
	year = {2022},
	note = {arXiv:2206.08325 [cs]},
}

@misc{mazeika_harmbench_2024,
	title = {{HarmBench}: {A} {Standardized} {Evaluation} {Framework} for {Automated} {Red} {Teaming} and {Robust} {Refusal}},
	shorttitle = {{HarmBench}},
	url = {http://arxiv.org/abs/2402.04249},
	doi = {10.48550/arXiv.2402.04249},
	urldate = {2026-04-25},
	publisher = {arXiv},
	author = {Mazeika, Mantas and Phan, Long and Yin, Xuwang and Zou, Andy and Wang, Zifan and Mu, Norman and Sakhaee, Elham and Li, Nathaniel and Basart, Steven and Li, Bo and Forsyth, David and Hendrycks, Dan},
	month = feb,
	year = {2024},
	note = {arXiv:2402.04249 [cs]},
}

@misc{pandey_socialharmbench_2026,
	title = {{SocialHarmBench}: {Revealing} {LLM} {Vulnerabilities} to {Socially} {Harmful} {Requests}},
	shorttitle = {{SocialHarmBench}},
	url = {http://arxiv.org/abs/2510.04891},
	doi = {10.48550/arXiv.2510.04891},
	urldate = {2026-04-25},
	publisher = {arXiv},
	author = {Pandey, Punya Syon and Le, Hai Son and Bhardwaj, Devansh and Mihalcea, Rada and Jin, Zhijing},
	month = feb,
	year = {2026},
	note = {arXiv:2510.04891 [cs]},
}

@misc{zeng_shieldgemma_2024,
	title = {{ShieldGemma}: {Generative} {AI} {Content} {Moderation} {Based} on {Gemma}},
	shorttitle = {{ShieldGemma}},
	url = {http://arxiv.org/abs/2407.21772},
	doi = {10.48550/arXiv.2407.21772},
	urldate = {2026-04-25},
	publisher = {arXiv},
	author = {Zeng, Wenjun and Liu, Yuchi and Mullins, Ryan and Peran, Ludovic and Fernandez, Joe and Harkous, Hamza and Narasimhan, Karthik and Proud, Drew and Kumar, Piyush and Radharapu, Bhaktipriya and Sturman, Olivia and Wahltinez, Oscar},
	month = aug,
	year = {2024},
	note = {arXiv:2407.21772 [cs]},
}

@inproceedings{cui_or-bench_2025,
	title = {{OR}-{Bench}: {An} {Over}-{Refusal} {Benchmark} for {Large} {Language} {Models}},
	shorttitle = {{OR}-{Bench}},
	url = {https://openreview.net/forum?id=CdFnEu0JZV},
	booktitle = {Proceedings of the 42nd International Conference on Machine Learning},
	language = {en},
	urldate = {2026-04-25},
	author = {Cui, Justin and Chiang, Wei-Lin and Stoica, Ion and Hsieh, Cho-Jui},
	month = jun,
	year = {2025},
}

@misc{zhang_falsereject_2025,
	title = {{FalseReject}: {A} {Resource} for {Improving} {Contextual} {Safety} and {Mitigating} {Over}-{Refusals} in {LLMs} via {Structured} {Reasoning}},
	shorttitle = {{FalseReject}},
	url = {http://arxiv.org/abs/2505.08054},
	doi = {10.48550/arXiv.2505.08054},
	language = {en},
	urldate = {2026-04-25},
	publisher = {arXiv},
	author = {Zhang, Zhehao and Xu, Weijie and Wu, Fanyou and Reddy, Chandan K.},
	month = jul,
	year = {2025},
	note = {arXiv:2505.08054 [cs]},
}

@inproceedings{sullutrone_cover_2025,
	address = {Vienna, Austria},
	title = {{COVER}: {Context}-{Driven} {Over}-{Refusal} {Verification} in {LLMs}},
	isbn = {979-8-89176-256-5},
	shorttitle = {{COVER}},
	url = {https://aclanthology.org/2025.findings-acl.1243/},
	doi = {10.18653/v1/2025.findings-acl.1243},
	urldate = {2026-04-25},
	booktitle = {Findings of the {Association} for {Computational} {Linguistics}: {ACL} 2025},
	publisher = {Association for Computational Linguistics},
	author = {Sullutrone, Giovanni and Vigliermo, Riccardo A. and Bergamaschi, Sonia and Sala, Luca},
	editor = {Che, Wanxiang and Nabende, Joyce and Shutova, Ekaterina and Pilehvar, Mohammad Taher},
	month = jul,
	year = {2025},
	pages = {24214--24229},
}

@techreport{noauthor_international_nodate,
	author = {{International AI Safety Report}},
	title = {International {AI} {Safety} {Report} 2026},
	institution = {International AI Safety Report},
	month = feb,
	year = {2026},
	url = {https://internationalaisafetyreport.org/publication/international-ai-safety-report-2026},
	urldate = {2026-04-25},
}

\appendix

\section{Systems Measurement Details}\label{app:systems_measurement}

We measure runtime on a single NVIDIA RTX 6000 Ada Generation GPU with 48GB memory, driver version 575.57.08, CUDA 12.8, PyTorch 2.11.0, and Transformers 5.5.4. All latency measurements use greedy decoding with \texttt{Qwen/Qwen3{-}8B}, \texttt{disable\_thinking}, a maximum of 64 generated answer tokens, and the same assistant system prompt used in the prompt-trigger experiments. Times include generator decoding and monitor or guard computation, but not model-loading time.

The harmful-prompt latency benchmark samples the same prompt families used in Table~\ref{tab:prompt_trigger_results}: 32 HarmBench DirectRequest prompts and 31 SocialHarmBench prompts. For \textsc{AERIC}, the monitor reads hidden states already produced by the generator and computes the same-pass heads during decoding. For WildGuard Prefix and ShieldGemma{-}9B Prefix, we call the guard every 8 emitted answer tokens on the current prompt-response prefix. For Qwen3Guard{-}Stream{-}4B, we update the native streaming guard with each newly emitted guard-token delta. The measurement is fixed-generation rather than stop-policy latency: generation continues up to the same 64-token cap, so the comparison isolates monitoring overhead rather than savings from early stopping.

\begin{table*}[h]
\centering
\scriptsize
\setlength{\tabcolsep}{4pt}
\begin{tabular}{llrrrr}
\toprule
Dataset & Method & Rows & Mean latency (ms) & P95 latency (ms) & Overhead \\
\midrule
HarmBench DirectRequest & No-stop generation & 32 & 1209.17 & 1437.83 & 0.00\% \\
HarmBench DirectRequest & \textsc{AERIC} same-pass & 32 & 1237.94 & 1482.93 & 2.38\% \\
HarmBench DirectRequest & Qwen3Guard{-}Stream{-}4B & 32 & 2149.33 & 2593.19 & 77.61\% \\
HarmBench DirectRequest & ShieldGemma{-}9B Prefix & 32 & 3136.53 & 3904.63 & 157.68\% \\
HarmBench DirectRequest & WildGuard Prefix & 32 & 3855.81 & 4713.28 & 218.88\% \\
\addlinespace[2pt]
SocialHarmBench & No-stop generation & 31 & 1116.18 & 1461.56 & 0.00\% \\
SocialHarmBench & \textsc{AERIC} same-pass & 31 & 1141.91 & 1494.02 & 2.31\% \\
SocialHarmBench & Qwen3Guard{-}Stream{-}4B & 31 & 2024.23 & 2662.08 & 81.41\% \\
SocialHarmBench & ShieldGemma{-}9B Prefix & 31 & 2912.53 & 3978.91 & 159.91\% \\
SocialHarmBench & WildGuard Prefix & 31 & 3508.75 & 4756.30 & 214.35\% \\
\bottomrule
\end{tabular}
\caption{Absolute latency and overhead for the harmful-prompt fixed-generation benchmark. Overhead is computed relative to the no-stop generation baseline measured on the same dataset and run configuration.}\label{tab:systems_latency_appendix}
\end{table*}

Aggregating the two prompt families by row count gives mean total latency 1190.69 ms for \textsc{AERIC}, 2087.77 ms for Qwen3Guard{-}Stream{-}4B, 3026.31 ms for ShieldGemma{-}9B Prefix, and 3685.04 ms for WildGuard Prefix. The corresponding overheads over no-stop generation are 2.34\%, 79.40\%, 158.73\%, and 216.74\%, respectively.

\section{Calibration and Threshold Details}\label{app:calibration_details}

The balanced response-classification benchmarks do not use a tuned decision threshold in the headline table. We report AUROC and AUPRC from the final EMA score, so those results should be read as threshold-free ranking evaluations.

For prompt-level trigger benchmarks, \textsc{AERIC} uses a source-side safe-budget rule. Let \(\mathrm{SafeTrigger}(\tau)\) denote the fraction of safe source-calibration continuations that would trigger at threshold \(\tau\), and let \(\mathrm{HarmTrigger@}K(\tau)\) denote the fraction of harmful source-calibration continuations that trigger within the first \(K\) monitored answer tokens. We select
\begin{equation}
    \tau^\star =
    \arg\max_{\tau:\ \mathrm{SafeTrigger}(\tau) \le B}
    \mathrm{HarmTrigger@}K(\tau),
\end{equation}
with \(B=0.10\) and \(K=16\) in the main prompt-trigger table. The threshold is chosen before evaluating HarmBench DirectRequest or SocialHarmBench.

\begin{table*}[h]
\centering
\scriptsize
\setlength{\tabcolsep}{4pt}
\begin{tabular}{lrrrr}
\toprule
Generator & Safe budget & Threshold & Source safe trigger & Source harm@16 \\
\midrule
\texttt{Qwen3{-}8B} & 5\% & 2.3861 & 0.0480 & 0.7656 \\
\texttt{Qwen3{-}8B} & 10\% & 1.3938 & 0.1000 & 0.8984 \\
\texttt{Gemma{-}4{-}E4B{-}it} & 5\% & 3.5613 & 0.0480 & 0.6797 \\
\texttt{Gemma{-}4{-}E4B{-}it} & 10\% & 2.6571 & 0.0947 & 0.7891 \\
\bottomrule
\end{tabular}
\caption{Source-side safe-budget calibration statistics for prompt-level trigger experiments. The main table uses the 10\% budget rows.}\label{tab:appendix_calibration}
\end{table*}

\section{Monitor Size and Hyperparameters}\label{app:monitor_hyperparams}

The default \textsc{AERIC} monitor uses three linear heads after a 128-dimensional projection: future hazard, support counterevidence, and paired residual drift. Each head has 128 weights plus one bias, for \(3 \times 129 = 387\) trainable head parameters. The fixed projection matrices and normalization statistics are stored in the monitor artifact but are not additional trainable head parameters. In fp32, the stored monitor artifact contains about 1.60M scalars for \texttt{Qwen/Qwen3{-}8B} (6.1 MiB) and about 1.00M scalars for \texttt{google/gemma{-}4{-}E4B{-}it} (3.8 MiB).

\begin{table}[t]
\centering
\scriptsize
\setlength{\tabcolsep}{4pt}
\begin{tabular}{ll}
\toprule
Hyperparameter & Value or grid \\
\midrule
Hidden layer & \(-8\) \\
Projection dimension & 128 \\
Future horizon \(H\) & 16 tokens \\
Linear probe regularization \(C\) & 0.1 \\
FineHarm train/calibration/test cap & 1024 / 256 / 256 rows \\
Paired residual source & 1024 safe + 1024 unsafe HH rows \\
Matched prompt contexts & 981 \\
Support weights \(\alpha\) & \(\{0.5, 1.0, 2.0\}\) \\
Residual weights \(\beta\) & \(\{0.0, 0.25, 0.5, 1.0, 2.0\}\) \\
Residual tail fraction & \(\{0.5, 1.0\}\) \\
EMA smoothing \(\lambda\) & 0.3 fixed \\
\bottomrule
\end{tabular}
\caption{Main monitor hyperparameters. The score-composition weights \(\alpha\) and \(\beta\) are selected from the listed grid by development AUROC\@; \(\lambda\) is fixed rather than jointly tuned.}\label{tab:appendix_hyperparams}
\end{table}

To check whether the EMA smoother is sensitive to this choice, Table~\ref{tab:appendix_lambda_sensitivity} varies \(\lambda\) for the default \texttt{Qwen/Qwen3{-}8B} balanced-benchmark monitor while leaving the learned heads fixed. Balanced AUROC changes only modestly across the tested range: from 0.7196 to 0.7096 on DiaSafety and from 0.8596 to 0.8563 on Harmful Advice. We therefore do not observe a sharp optimum, and \(\lambda=0.3\) is retained as a middle operating point rather than as a brittle tuned value.

\begin{table}[t]
\centering
\scriptsize
\setlength{\tabcolsep}{4pt}
\begin{tabular}{lccccc}
\toprule
Dataset & \(\lambda=0.1\) & \(\lambda=0.2\) & \(\lambda=0.3\) & \(\lambda=0.5\) & \(\lambda=0.7\) \\
\midrule
DiaSafety AUROC & \textbf{0.7196} & 0.7159 & 0.7143 & 0.7124 & 0.7096 \\
Harmful Advice AUROC & 0.8590 & \textbf{0.8596} & 0.8582 & 0.8565 & 0.8563 \\
\bottomrule
\end{tabular}
\caption{Sensitivity of the EMA smoothing coefficient \(\lambda\) for the default \texttt{Qwen/Qwen3{-}8B} balanced-benchmark monitor. The learned heads are held fixed and only the EMA smoothing parameter is changed at evaluation time. Performance is stable across the tested range, so we keep \(\lambda=0.3\) as a central default.}\label{tab:appendix_lambda_sensitivity}
\end{table}

\section{Baseline Protocols}\label{app:baseline_protocols}

For balanced response-classification benchmarks, post-hoc guards score completed prompt-response pairs. WildGuard is decoded deterministically and parsed into a harmful-response decision. ShieldGemma{-}9B is run over the prompt-response pair under four policy guidelines: dangerous content, harassment, hate speech, and sexually explicit information. We use the maximum unsafe probability across these categories as the ShieldGemma score.

For prompt-trigger benchmarks, all prefix baselines are replayed over the same realized no-stop continuations used by \textsc{AERIC}. WildGuard Prefix and ShieldGemma{-}9B Prefix are called every 8 emitted answer tokens on the current prompt-response prefix. ShieldGemma{-}9B Prefix again uses the maximum unsafe probability across the four guideline categories and triggers when the resulting unsafe decision is positive. Qwen3Guard{-}Stream{-}4B is evaluated as a native streaming guard: the prompt is first sent as the user stream, and each newly emitted assistant guard-token delta is then passed to the guard. The stream triggers when the latest native risk level is \texttt{Unsafe}.

The prompt-trigger comparison is therefore not a post-hoc full-output comparison. It asks when each method would first fire on the same deterministic generation path. This replay design avoids confounding guard quality with different generator samples.

\section{Dataset and Preprocessing Details}\label{app:dataset_preprocessing}

DiaSafety and Harmful Advice are treated as balanced implicit-harm response-classification benchmarks. DiaSafety contains 1094 test rows in our run; Harmful Advice contains 550 rows. The headline metrics for these datasets are AUROC and AUPRC over completed response trajectories.

HarmBench DirectRequest and SocialHarmBench are treated as prompt-only harmful request suites. HarmBench DirectRequest contains 320 prompts. SocialHarmBench contains 585 raw prompts in the loaded split; one prompt is empty after field extraction in our run, yielding 584 non-empty prompts for the full prompt-trigger table. These datasets are not used as balanced response-classification benchmarks in the paper.

The prompt-trigger generation setup uses greedy decoding, a maximum answer length of 128 tokens for the main trigger table, and a system prompt instructing the generator to provide direct answers without refusal or safety disclaimers. The latency appendix uses the same prompt families but caps generation at 64 tokens to isolate fixed-generation monitoring overhead.

\section{Uncertainty Interval Details}\label{app:uncertainty_intervals}

For the balanced implicit-harm benchmarks in Table~\ref{tab:balanced_results}, the bracketed intervals are nonparametric $95\%$ bootstrap confidence intervals. For each method, benchmark, and metric, we repeatedly resample evaluation rows with replacement from the fixed test set, recompute the metric on each bootstrap sample, and report the percentile interval. The source of variability is therefore the finite evaluation sample rather than retraining or random-seed variation: all generators, guards, and \textsc{AERIC} monitors are kept fixed while the test rows are resampled.

\end{document}